Accepted Version - *Journal of Thermal Biology*, december 2022.
The publisher version can be found with the DOI: 10.1016/j.jtherbio.2023.103464

# Facial Thermal and Blood Perfusion Patterns of Human Emotions: Proof-of-Concept

Victor H. Aristizabal-Tique[a], Marcela Henao-Pérez[b], Diana Carolina López-Medina[b], Renato Zambrano-Cruz[c] and Gloria Díaz-Londoño[d*]

[a]*School of Engineering, Universidad Cooperativa de Colombia, Medellín, Colombia; 050012. vharisti@yahoo.com, victor.aristizabalt@campusucc.edu.co*

[b]*School of Medicine, Universidad Cooperativa de Colombia, Medellín, Colombia; 050012. marcela.henaop@campusucc.edu.co, diana.lopezme@campusucc.edu.co*

[c]*School of Psychology, Universidad Cooperativa de Colombia, Medellín, Colombia; 050012. renato.zambrano@ucc.edu.co*

[d]*School of Science, Universidad Nacional de Colombia-Sede Medellín, Colombia; 050034. gmdiazl@gmail.com, gmdiazl@unal.edu.co*

*Corresponding author: gmdiazl@unal.edu.co

**Abstract:** In this work, a preliminary study of proof-of-concept was conducted to evaluate the performance of the thermographic and blood perfusion data when emotions of positive and negative valence are applied, where the blood perfusion data are obtained from the thermographic data. The images were obtained for baseline, positive, and negative valence according to the protocol of the Geneva Affective Picture Database. Absolute and percentage differences of average values of the data between the valences and the baseline were calculated for different regions of interest (forehead, periorbital eyes, cheeks, nose and upper lips). For negative valence, a decrease in temperature and blood perfusion was observed in the regions of interest, and the effect was greater on the left side than on the right side. In positive valence, the temperature and blood perfusion increased in some cases, showing a complex pattern. The temperature and perfusion of the nose was reduced for both valences, which is indicative of the arousal dimension. The blood perfusion images were found to be greater contrast; the percentage differences in the blood perfusion images are greater than those obtained in thermographic images. Moreover, the blood perfusion images, and vasomotor answer are consistent, therefore, they can be a better biomarker than thermographic analysis in identifying emotions.

## 1 Introduction

Emotions are psychological processes considered as responses to stimuli that involve brain processes, body reactions, and behavior [1], [2]. The response to emotions comprises multidimensional processes of short duration, is subjective, has a high affective load, and seeks balance or homeostasis [3]. The autonomic nervous system generates different types of physiological and behavioral responses that are important markers, such as, an individual’s pallor, blush, or pupil size, which associate with facial expressions, and allow the nonverbal communication of emotions [2]. Both the skin and the muscles of the face and the blood vessels that irrigate them are controlled by the action of this autonomous system [4].

There are three main approaches in the study of emotions: the modular (or discrete), the dimensional, and the one that combines elements of both the modular and dimensional [5]. The modular approach considers that emotions can be classified in six basic categories: happiness, surprise, disgust, anger, fear, and sadness. The dimensional approach considers that emotions can be classified based on the dimensions of valence that identify if the emotion is pleasant-unpleasant and arousal, which measures the intensity of the low-high emotion or the activation-deactivation [6]. The third approach is based on the combination of a dimensional and a modular approach. Figure 1 shows the classification at the level of valence and arousal; yellow represents the emotion associated with the modular approach for happiness, disgust, anger, fear, and sadness. Specifically, it can be observed that surprise is an emotion of neutral valence and high activation [6].

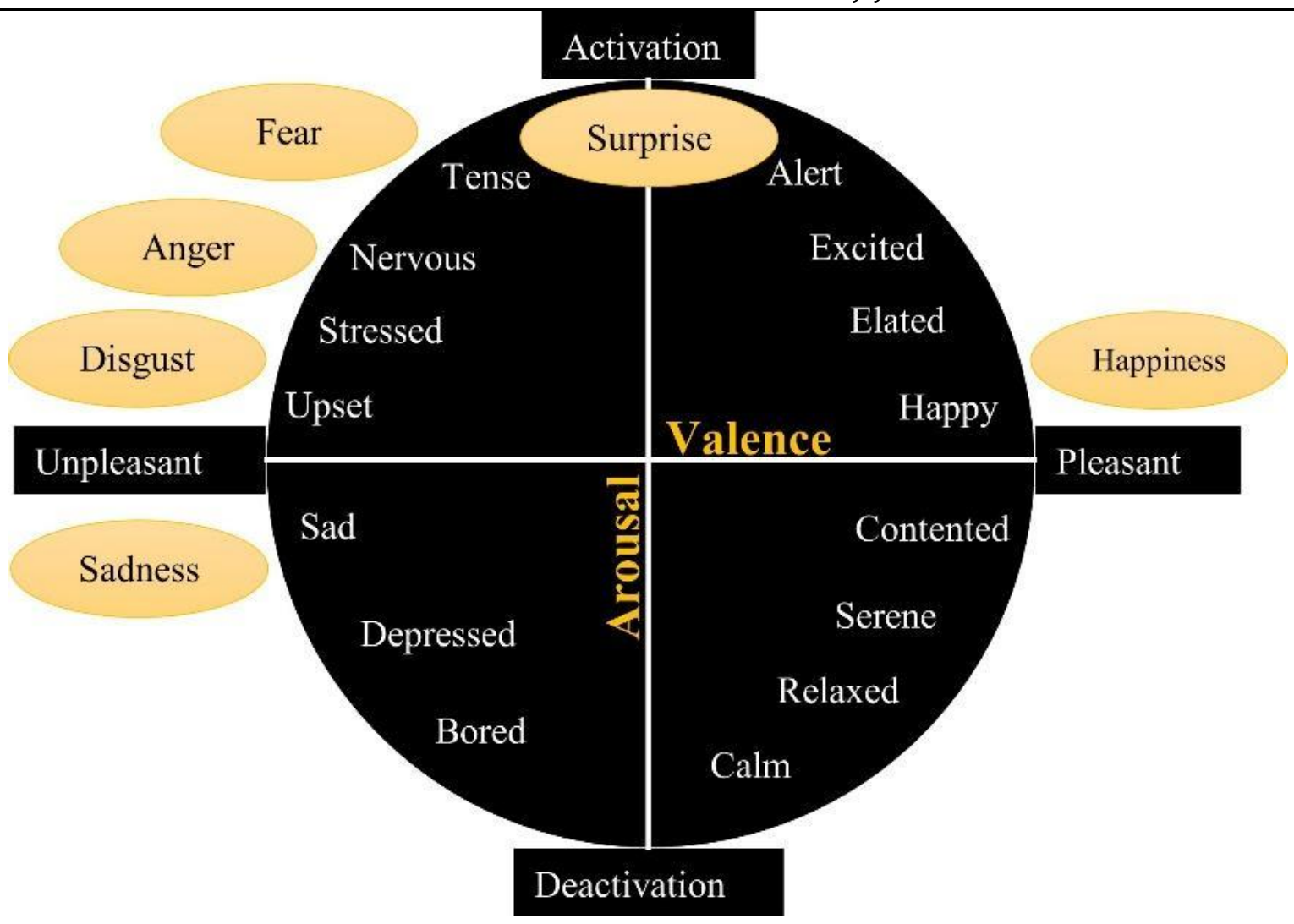

**Figure 1.** Classification of valence and arousal in emotions. Yellow shows the classification of the modular approach. Circumplexed model of emotions, adapted from [6].

The approaches mentioned above have different evaluation strategies. On the one hand, some are focused on self-reporting measures, whereas others focus on the analysis of expected nonverbal behavior. On the other hand, some pay attention to the chemical analysis of hormones or neurotransmitters, while others attend to the physiological analysis of aspects such as heart rate, galvanic skin response, or respiratory rate [7]. All these evaluation methods allow one to establish a dispositional state for behavior and decision making so that predictions can be generated in different fields, such as marketing, education, or health [7].

Some previous studies have evaluated the modular approach to emotions based on facial expression [8], [9]. Because emotions generate micro expressions in the face, identifying the muscle involved in the movement can help determine the emotion [10]. In these works, the authors have used images where the face is visible, captured by traditional RGB (Red, Green, and Blue) cameras. In some cases, algorithms for the automatic identification of emotions have been implemented [11]; moreover, in recent years, artificial intelligence has been used for this purpose [12].

To evaluate the dimensional approach and the combination of the two approaches (modular and dimensional), studies of emotions have been developed based on infrared thermography, where the thermal changes in the face that can be associated with an emotional response [13]–[17], which has gained considerable interest as a non-invasive technique. In this method, infrared radiation from the skin surface is captured and associated with blood perfusion changes [2], [17], [18], moreover, it is a low-noise measurement [18].

In several studies on infrared thermography, it has been found that thermal changes in the tip of the nose allow one to estimate the emotional activation or arousal dimension because there are no underlying muscles in this region; therefore, thermal changes due to muscle contraction are avoided [2], [17]. This has been demonstrated in studies on primates such as the work carried out by Kuraoka and Nakamura [19]. In the case of humans, a temperature decrease on the nose tip is an indication of the arousal measurement has been reported [2], [19]–[22].

Previous works [14], [16], [17] have shown that analyzing several points in the face, such as the cheeks, periorbital region, upper lip region, and forehead, can allow one to identify the type of emotion defined in the modular approach by analyzing if there is an increase or decrease in temperature in those regions. Other earlier studies [21], [23], [24] have used pattern recognition techniques by considering characteristics from thermal images to classify emotions, for example, the work of Khan et al. [23] reported success rates between 56% and 66.28%; Nhan and Chau [21] reported success rates between 50% and 60% in high and low arousal and valence classification.

Conversely, earlier works perform facial recognition of individuals [25]–[29] based on images of blood perfusion patterns generated from thermographic scans. In this case, an algorithm supported by the heat transfer model defined by the Pennes equation [25], [27] is implemented. The blood perfusion images generated by this algorithm show a greater contrast and detail compared to the thermographic images. The authors have found success rates in facial recognition that are above 90% in controlled environmental conditions and approximately 80% in uncontrolled environments [25]. In the case of thermographic images, the success rate is approximately 60% in an uncontrolled environment, whereas it is approximately 80% in a controlled environment. This shows that the thermographic image is limited by environmental conditions [25].

In other works about recognition with images of blood perfusion, authors carried out acquisition of thermal images and have used the heat transfer algorithm to identify stress level [30], [31], sexual arousal [32], [33] and personality [34], [35]. Puri et al. [30], [31] observed a difference of forehead blood flow between baseline measurements and during a stressor (Stroop session). Basu et al. [34], [35] evaluated happiness, anger, surprise and disgust in the whole face, the authors acquired a thermal images of the face and converted them into corresponding images of blood perfusion; subsequently, the eigenfeatures are obtained from the blood perfusion image and visible image, both features are fused and after, the SVM (Support Vector Machine) is used as a classification technique to estimate the personality as Psychotic (aggressiveness and hostility), Extravert (energetic and social person) or Neurotic (unstable emotionality).

Considering previous research and the need to obtain a higher success rate in the identification of emotions with a higher degree of independence from environmental conditions, this work aims to propose a preliminary proof of concept study, through the case study methodology, to evaluate performance of the thermographic and blood perfusion data when emotions of positive and negative valence are applied, where the blood perfusion data are obtained from the thermographic data by means of the model proposed by Wu et al [25], [29]. Here, the blood perfusion algorithm works as a type of "band pass filter" that removes the ambient thermal noise, leaving only the physiological effect generated by the emotion, allowing so, a greater change and contrast in the facial pattern, resulting in a measurement proposal with a high degree of independence from environmental conditions.

Finally, we analyze the performance of the thermal and blood perfusion changes in different regions of interest (ROIs) of the face, in order to demonstrate that the blood perfusion changes estimated with the algorithm can be a better biomarker of response to emotions than thermal changes. Furthermore, these blood perfusion patterns are analyzed from a psychophysiological approach to see the consistency of the results.

## 2 Materials and Methods

In this work, a quantitative preliminary study of proof-of-concept is conducted using the case study methodology to evaluate the efficacy of the blood perfusion algorithm versus thermographic imaging in a small group of Colombian participants, to establish key parameters and initial values for our population. It should be noted that this study was developed in a laboratory environment and seeks to report the basic principles found when executing a prototype in healthy subjects, who do not show psychological or neuropsychiatric disorders, skin and metabolic diseases. Three subjects, two women and one man, aged 18–45, were tested. Each subject visualized images that were validated for the baseline emotions, both of positive and negative valence, according to the protocol described in section 2.1. Thermographic images were taken during the presentation of the images associated with each emotion as per section 2.2. Subsequently, the thermographic images were processed, according to section 2.3, applying the algorithm proposed by Wu et al [25] and thus obtaining the blood perfusion images of each individual's face. The average values of temperature and blood perfusion were calculated for every emotion. The ROIs analyzed were the nose, forehead, eyes, upper lip region, and total face evaluation, according to the recommendations of previous works [14], [16], [17]. For the quantitative analysis of the results, the absolute and percentage differences between the negative valence-baseline and positive valence-baseline images for temperature and blood perfusion were calculated.

### *2.1 Stimuli*

We selected 90 pictures (30 pleasant, 30 unpleasant, and 30 neutral) from the Geneva Affective Picture Database (GAPED) [36]. We took the GAPED pictures according to original values from the study of Dan-Glaser and Scherer, where the ratings were ranged from 0 to 100. The pictures used in this study were selected considering the most extreme values of the original study to guarantee the evocation of the affective state, therefore, these pictures have an averaged rating of 96 for the pleasant pictures, 15 for the unpleasant pictures, and 50 for the neutral pictures.

The set of pleasant pictures included landscapes, newborns, children playing, happy families, puppies, and domestic and wild animals in their habitats. The set of unpleasant images consisted of animals that were hunting or feeding; dead, slaughtered, or injured animals; reptiles; and arachnids; in the images that had people, there were corpses of adults, people in rites or in contact with war weapons, and images of infants in a state of extreme malnutrition. The set of neutral images presented supplies, work equipment, and clean and illuminated rooms without people; among these elements, there were electrical cables, bases, envelopes, stairs, chairs, furnished offices, bathrooms, and so on.

The stimulation protocol began with 15 minutes of space acclimatization. At the end of this period, a thermographic video was recorded for one minute as a baseline in order to start with the presentation of the stimuli for the following 18 minutes. Each set of images lasted four minutes. It started with those of negative valence, continued with those of neutral valence, and concluded with those of positive valence. Each image was transmitted for four seconds; this was followed by a black background image with a white plus sign in the center, which was also projected for four seconds. There was a three-minute break between each set. The stimulation protocol lasted 33 minutes, from the acclimatization period to the last image of the set of positive valence stimulus, as shown in Figure 2.

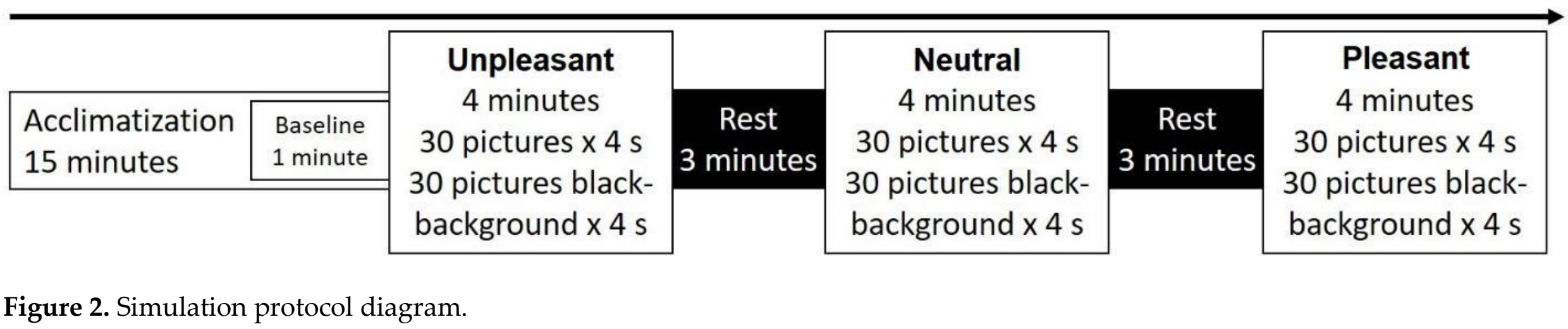

**Figure 2.** Simulation protocol diagram.

### *2.2. Protocol for the acquisition of thermographic images*

The measurements were taken in Cabin 5 of the Psychology Laboratory of Universidad Cooperativa de Colombia, the Medellín campus. The area of the cabin was approximately 15 m², and it had an air conditioner that allowed for the temperature and relative humidity of the room to be controlled. The room had no exposed heating pipes or electrical wiring because these elements could have interfered with the thermographic measurement, as they are sources of heat. The thermographic images were obtained using a FLIR A655SC thermographic camera with a wavelength in the far infrared band (7.5–14 μm), a spatial resolution of 640 × 480 pixels, an accuracy of ±2 °C, and a NETD (Noise Equivalent Temperature Difference, also known as thermal sensitivity) over 30 mK (0.03 °C); the software used for the acquisition was the FLIR R&D version 3.3 [18].

The distance between the FLIR A655SC camera and the subjects' faces was 0.80 m; the room temp was 24 °C; and the relative humidity was 63%, these factors were measured by hygro-thermometer. The measurements were taken in the morning because individuals' body temperatures fluctuate less at that time owing to the circadian cycle [37]. It was previously verified that none of the study subjects presented skin alterations or any diseases that could alter skin temperature. In addition, the subjects were instructed not to engage in physical activity or consume alcohol, coffee, or energy drinks during the 12 hours before the test to avoid any energy alterations. The subjects did not apply any creams or lotions on their faces to avoid alterations in the emissivity of the skin. Prior to the acquisition of the thermographic images and emotional stimulation, the subjects relaxed for 15 minutes (see Figure 2). In general, the protocol of the acquisition of thermographic imaging was in accordance with recommendations from previous studies [38], [39], and the setup for it can be observed in Figure 3 (a).

The subjects sat on chairs without cushions in order to decrease body movement and their faces were immobilized as shown in Figure 3 (b), and the height was adjusted to horizontally match, on an imaginary line, the center of the camera and the nose of each participant so that their entire face could be imaged. Finally, for each subject and emotion, thermographic videos were taken at a frequency of 1 Hz (one image per second); approximately 60 thermographic images were acquired for the baseline set, 240 for the positive valence set, and 240 for the negative valence set.

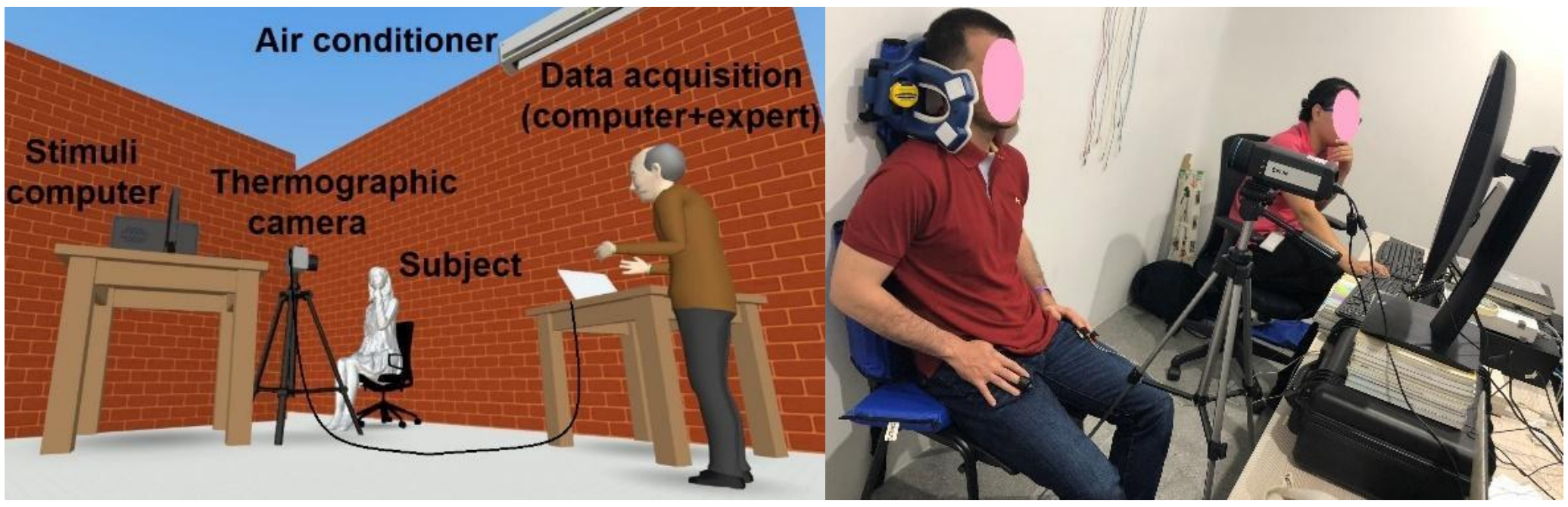

(a) (b)

**Figure 3. (a)** Experimental setup for the acquisition of thermographic data, **(b)** immobilization system for taking facial thermographic images.

### *2.3. Processing the thermographic images*

To process and analyze the images, a first code was implemented in MATLAB, which performs the Otsu thresholding to segment each image and obtain only the face of each subject studied. Subsequently, the algorithm proposed by Wu et al. [26], [27], [29] was applied to obtain the images of blood perfusion, as described in section 2.4. In the thermographic and perfusion images, ROIs of 10 × 10 pixels were defined for the nose, eyes, and upper lip; in the case of the forehead, the ROI was a 50 × 110 pixel rectangle, as shown in Figure 4.

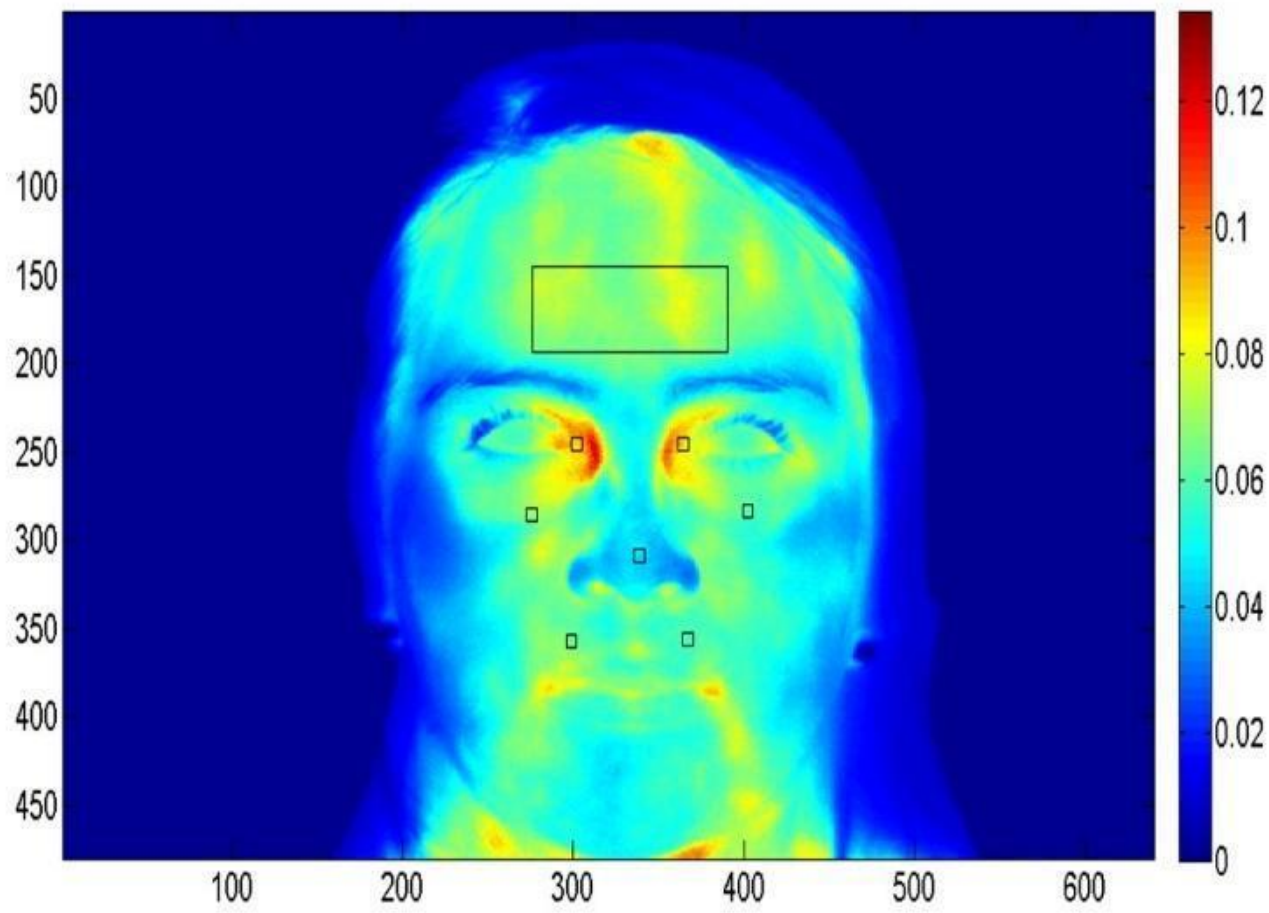

**Figure 4.** Regions of interest defined for image analysis (the nose, forehead, eyes and upper lip).

For each ROI, the average temperature and average blood perfusion were determined. Additionally, the average temperature and average blood perfusion of the entire face were calculated. For each ROI, six output files of the average temperature and blood perfusion associated with the thermographic videos obtained during the baseline time periods, negative valence (fear), and positive valence (happiness) were generated. Of these files, two have 60 elements corresponding to the baseline set, two have 240 data corresponding to the negative valence set, and the last two also have 240 data corresponding to the positive valence set. Figure 5 presents a summary diagram of the processing of each of the thermographic images of the video.

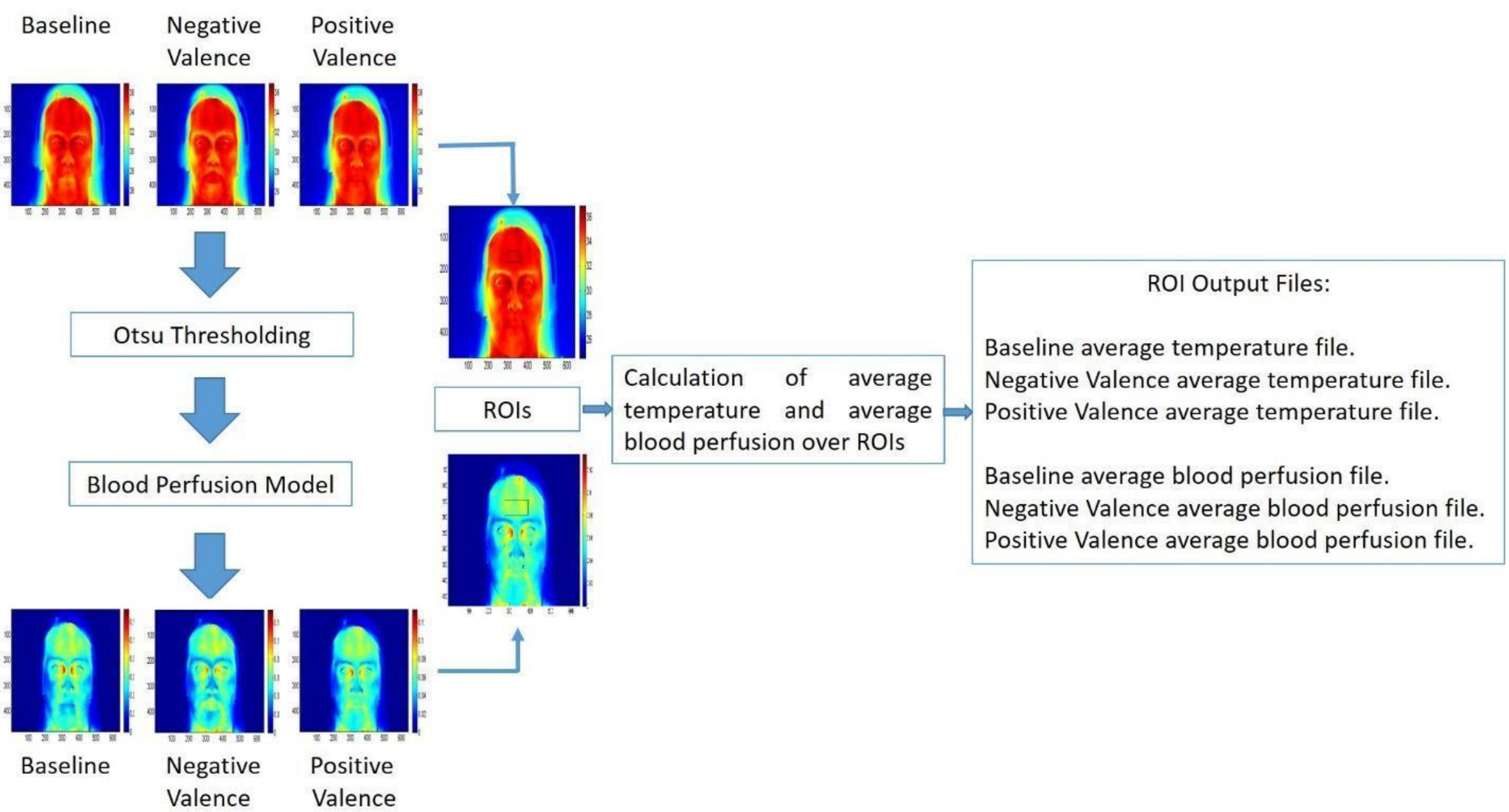

**Figure 5.** Processing diagram of each image of the thermographic video.

These output files were read by a second code in MATLAB that calculated the average temperature or perfusion value of the 60 or 240 data, depending on the case of the specific set analyzed (baseline, negative valence, or positive valence) and the ROI. Subsequently, the values of the absolute differences and percentage between the average temperatures of the negative valence-baseline and positive valence-baseline were calculated as follows:

$$\Delta T_{\alpha B} = T_\alpha - T_B \quad (1)$$

$$\Delta T_{\alpha B}\% = \frac{(T_\alpha - T_B)}{T_B} \times 100\% \quad (2)$$

where $T_\alpha$ is the average temperature for negative valence (Fear, $\alpha$=F) or positive valence (Happinness, $\alpha$=H) and $T_B$ is the average temperature of the baseline. Thus, the absolute and percentage differences between the average blood perfusion of the negative valence-baseline and the positive valence-baseline were calculated as follows:

$$\Delta \omega_{\alpha B} = \omega_\alpha - \omega_B \quad (3)$$

$$\Delta \omega_{\alpha B}\% = \frac{(\omega_\alpha - \omega_B)}{\omega_B} \times 100\% \quad (4)$$

where $\omega_\alpha$ represents the average blood perfusion for negative valence (Fear, $\alpha$=F) or positive valence (Happinness, $\alpha$=H) and $\omega_B$ is the blood perfusion of the baseline. An outline of the quantitative analysis of the results is shown in Figure 6.

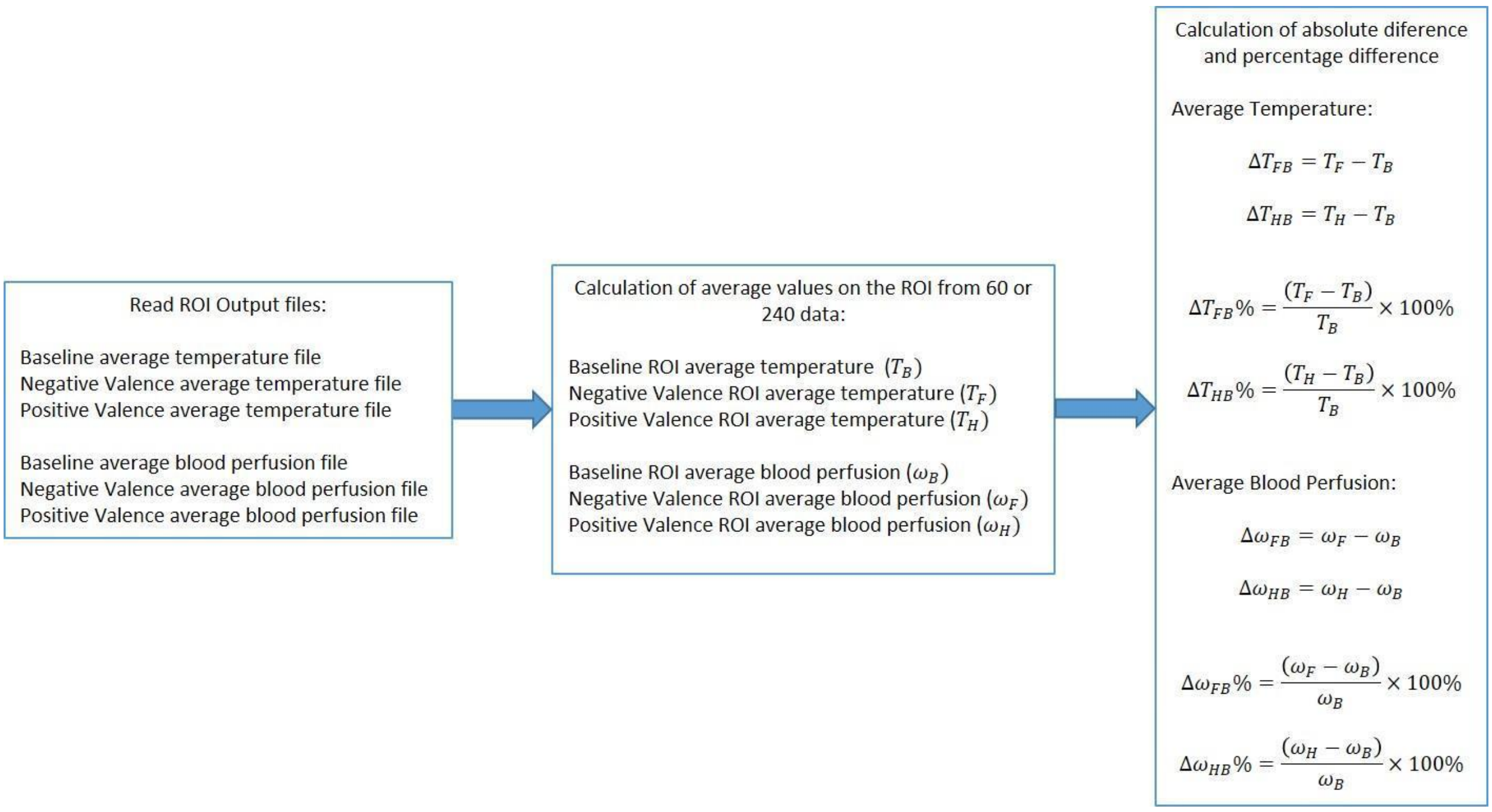

**Figure 6.** Diagram of the quantitative analysis of the average temperature and average blood perfusion results for each of the emotions and regions of interest analyzed.

### *2.4. Blood perfusion model*

The blood perfusion model is based on the heat transfer equation, which relates the loss of heat from the skin surface to the environment and heat production, which is caused by the added metabolism plus the conduction of heat from the center of the body to the surface of the skin, as transported by the blood [40]. The above is summarized in the following equation [25]–[27], [29]:

$$H_r + H_f + H_e = H_m + H_c + H_b \quad (5)$$

where $H_r$, $H_f$, and $H_e$, are the heat flows per unit area of radiation, convection, and evaporation, respectively. The radiation model is defined by the Stefan–Boltzmann law [25], [41] and the convection in air by Newton's law [25], [41]. The equations are as follows:

$$H_r = \varepsilon\sigma(T_s^4 - T_e^4) \quad (6)$$

$$H_f = AK_f d^{3M-1}\left(\frac{P_r g\beta}{\nu^2}\right)^M (T_s - T_e)^{M+1} \quad (7)$$

where $T_s$ is the temperature of the skin and $T_e$ is the ambient temperature. Because the image acquisition was carried out in a controlled environment, the heat flow per unit area due to evaporation is negligible, that is, $H_e \approx 0$. $H_m$, $H_c$, and $H_b$ correspond to the heat flow per unit area because of the metabolism, the heat conduction from the center of the body to the skin surface, and the convection of the blood perfusion, respectively. The value corresponding to the metabolism was taken from Pennes' model [25], [42] and is considered as follows: $H_m$= 4.186 w/m$^2$. The pattern of conduction from the inside of the body to the surface of the skin is given by Fourier's law [41] while that of blood perfusion convection is given by the Pennes model [25], [42]. The equations are as follows:

$$H_c = \frac{k_s(T_c - T_s)}{D} \quad (8)$$

$$H_b = \omega\alpha c_b(T_a - T_s) \quad (9)$$

where ω is the blood perfusion value (kg/m$^2$s), $T_s$ is the skin temperature acquired from the thermographic image, and $T_c$ is the central temperature, which is considered equal to the arterial temperature $T_a$ ($T_c$=$T_a$). The rest of the parameters of the equations with their respective values were taken from the work of Wu et al. [25] and are shown in Table 1.

Finally, by replacing Equations (6)–(9) in the heat transfer model (see Equation (5)) and clearing blood perfusion ω, the following expression is obtained:

$$\omega = \frac{\varepsilon\sigma(T_s^4 - T_e^4) + AK_f d^{3M-1}\left(\frac{Prg\beta}{\nu^2}\right)^M (T_s - T_e)^{M+1} - \frac{k_s(T_a - T_s)}{D} - 4.186}{\alpha c_b(T_a - T_s)} \quad (10)$$

**Table 1.** Values of the parameters of Equation (10). Taken from Wu et al. [25].

| Symbol | Representation | Value |
|---|---|---|
| $\rho_b$ | Blood density | 1060 kg/m$^3$ |
| $c_b$ | Blood specific heat | 3.78×10$^3$ J/(kg K) |
| $T_a$ | Arterial/core body temperature | 312.15 K |
| $T_c$ | Core body temperature | 312.15 K |
| $k_s$ | Tissue/skin thermal conductivity | 0.5 W/(m K) |
| $k_f$ | Air thermal conductivity | 0.024 W/(m K) |
| $Q_m$ | Metabolic heat flux per unit area | 4.186 W m$^{-2}$ |
| $\sigma$ | Stefan–Boltzmann constant | 5.67×10$^{-8}$ W/(m$^2$ K$^4$) |
| $\varepsilon$ | Tissue/skin thermal emissivity | 0.98 |
| $\alpha$ | Tissue/skin countercurrent exchange ratio | 0.8 |
| $P_r$ | Prandtl number | 0.72 |
| $\nu$ | Kinematic viscosity of air | 1.56×10$^{-5}$ m$^2$/s |
| $\beta$ | Air thermal expansion coefficient | 3.354×10$^{-3}$ K$^{-1}$ |
| $g$ | Local gravitational acceleration | 9.8 m$^2$/s |
| $A$ | Constant | 0.27 |
| $M$ | Constant | 0.25 |
| $d$ | The characteristic length of the object | 0.170 |
| $D$ | The distance from the body core to the skin surface | 0.085 m |

### *2.5. Statistic analysis*

A statistical validation of thermographic and blood perfusion changes for each subject and ROI was performed with repeated measures (within-subjects) analysis of variance (ANOVA) [43], [44], where the baseline values are the initial variable, and the measurements given either by the stimuli of positive or negative valence are the subsequent variable; therefore, the statistical analysis used follows the indications for dependent samples [45]. The statistical program used was Jamovi version 1.6, and a p value less than 0.05 was set as statistically significant [46].

Initially, the normality distribution of the data was determined using the Kolmogorov-Smirnov test since there were more than 200 data per variable. Subsequently, to determine the thermographic and blood perfusion changes after applying the stimuli of positive or negative valence with respect to the baseline state, parametric and non-parametric parameters were used (Student's t of related samples and Wilcoxon Rank test, respectively) [47].

## 3. Results and discussion from the performance of the thermographic and blood perfusion analysis

As indicated above, in this work, thermographic images were acquired for the baseline and of both negative and positive valence states. These images were processed using the blood perfusion algorithm, and a new set of images was thus obtained. Figure 7 shows the thermographic and blood perfusion of a woman for each of the states studied.

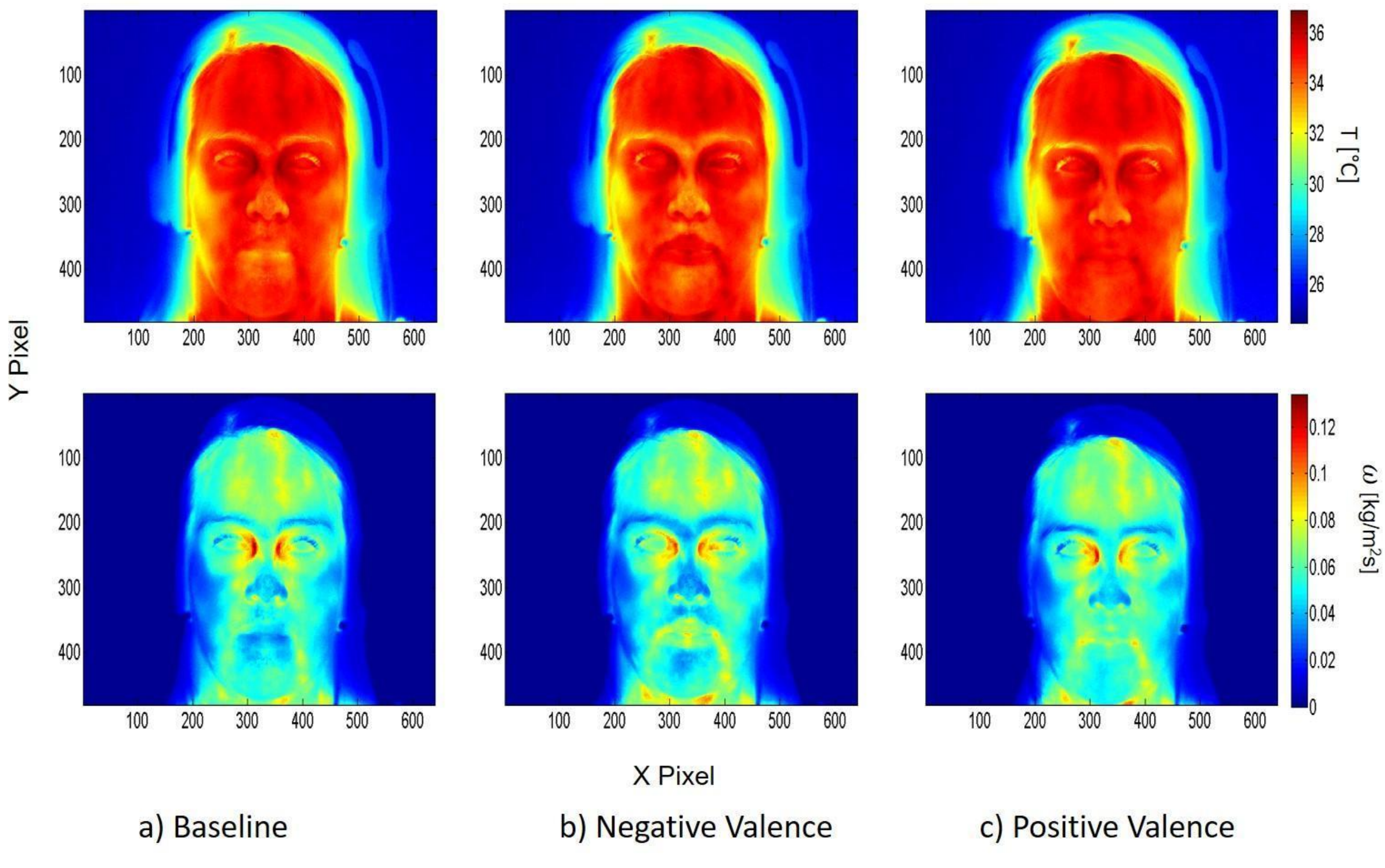

**Figure 7.** The upper panel shows the thermographic images while the lower panel shows the blood perfusion images of a woman in each state: (**a**) baseline, (**b**) negative valence (fear), and (**c**) positive valence (happiness).

When visualizing the thermographic images in Figure 7, a change in the nose region is observed for negative valence emotions and positive valence emotions compared to the baseline. In the other points of the face, it is difficult to observe a change at first sight; in this case, a quantitative analysis of the image is required. When the blood perfusion images are analyzed, the change is found to be significant for the emotions of negative valence and positive valence compared to the baseline in the nose, mouth, chin, and forehead regions. At first sight, the change between the blood perfusion images for each emotion is clear, and there is a better contrast and a greater level of detail than in thermographic images. Our result is consistent with what was reported in previous works [25]–[27].

### *3.1. Quantitative analysis of thermographic images*

In Table 2, the results of the average temperatures obtained for the baseline ($T_B$), negative valence ($T_F$), and positive valence ($T_H$) are presented for the case of the man and the two women. In addition, the absolute ($\Delta T_{FB}$ and $\Delta T_{HB}$) and percentage ($\Delta T_{FB}$% and $\Delta T_{HB}$%) differences, calculated with Equations (1) and (2), respectively, are presented.

**Table 2.** Average human temperature values in the each of regions of the interest for the baseline, negative valence, and positive valence and their absolute and percentage differences. Uncertainties are standard deviations, at 1 level, are given between parentheses.

| | $T_B\,[°C]$ **Baseline** | $T_F\,[°C]$ **Negative valence** | $T_H\,[°C]$ **Positive valence** | $\Delta T_{FB}\,[°C]$ | $\Delta T_{FB}$**%** | $\Delta T_{HB}\,[°C]$ | $\Delta T_{HB}$**%** |
|---|---|---|---|---|---|---|---|
| **Man 1** | | | | | | | |
| Nose-ǂ | 34.46 (0.14) | 34.18 (0.12) | 33.91 (0.20) | -0.28** | **-0.82** | -0.55** | **-1.59** |
| Forehead | 34.83 (0.02) | 34.76 (0.04) | 34.87 (0.02) | -0.07** | **-0.20** | 0.04 ** | **0.11** |
| Left eyeǂ | 35.73 (0.14) | 35.67 (0.14) | 35.81 (0.11) | -0.06 | **-0.17** | 0.08** | **0.22** |
| Right eyeǂ | 35.66 (0.28) | 35.57 (0.26) | 35.78 (0.12) | -0.10 | **-0.28** | 0.12** | **0.33** |
| Left cheek | 34.44 (0.12) | 34.25 (0.12) | 34.71 (0.06) | -0.19 ** | **-0.55** | 0.27 * | **0.78** |
| Right cheek | 34.82 (0.06) | 34.69 (0.09) | 34.81 (0.03) | -0.13 ** | **-0.37** | -0.01 ** | **-0.03** |
| Left upper lip | 34.70 (0.06) | 34.75 (0.04) | 34.70 (0.09) | 0.05 | **0.14** | 0.00 | **0.00** |
| Right upper lip | 34.52 (0.03) | 34.50 (0.06) | 34.52 (0.03) | -0.02** | **-0.06** | 0.00** | **0.00** |
| Total face | 33.81 (0.02) | 33.74 (0.04) | 33.77 (0.02) | -0.07** | **-0.21** | -0.05** | **-0.14** |
| **Woman 1** | | | | | | | |
| Noseǂ | 34.19 (0.35) | 33.84 (0.24) | 34.19 (0.21) | -0.34** | **-0.99** | 0.00 | **0.00** |
| Forehead | 35.52 (0.04) | 35.46 (0.10) | 35.47 (0.04) | -0.06**ǂ | **-0.17** | -0.05** | **-0.14** |
| Left eye | 36.11 (0.47) | 36.11 (0.19) | 35.98 (0.09) | 0.00 | **0.00** | -0.13 | **-0.36** |
| Right eyeǂ | 35.87 (0.88) | 36.02 (0.38) | 36.03 (0.19) | 0.14** | **0.39** | 0.16** | **0.44** |
| Left cheekǂ | 35.24 (0.17) | 35.26 (0.10) | 35.09 (0.06) | 0.02 | **0.06** | -0.15** | **-0.42** |
| Right cheekǂ | 35.47 (0.17) | 35.46 (0.09) | 35.49 (0.06) | -0.01 | **0.03** | 0.02** | **0.06** |
| Left upper lipǂ | 34.55 (0.17) | 34.40 (0.26) | 35.06 (0.09) | -0.15** | **-0.43** | 0.51** | **1.48** |
| Right upper lipǂ | 35.12 (0.37) | 34.81 (0.24) | 35.35 (0.09) | -0.31** | **-0.88** | 0.23** | **0.65** |
| Total face | 33.73 (0.05) | 33.73 (0.07) | 33.75 (0.04) | 0.00 | **0.00** | 0.02** | **0.06** |
| **Woman 2** | | | | | | | |
| Nose | 34.61 (0.22) | 34.41 (0.18) | 34.46 (0.29) | -0.20** | **-0.58** | -0.15**ǂ | **-0.43** |
| Forehead | 35.09 (0.04) | 34.83 (0.02) | 34.77 (0.03) | -0.26** | **-0.74** | -0.32** | **-0.91** |
| Left eyeǂ | 36.17 (0.06) | 35.83 (0.11) | 35.69 (0.18) | -0.34** | **-0.94** | -0.48** | **-1.33** |
| Right eyeǂ | 35.87 (0.24) | 35.52 (0.14) | 35.46 (0.40) | -0.36** | **-1.00** | -0.41** | **-1.14** |
| Left cheek | 34.38 (0.22) | 34.14 (0.26) | 34.24 (0.26) | -0.23** | **-0.67** | -0.13**ǂ | **-0.40** |
| Right cheek | 34.84 (0.17) | 34.69 (0.11) | 34.48 (0.41) | -0.15** | **-0.43** | -0.36** | **-1.04** |
| Left upper lip | 35.01 (0.12) | 34.86 (0.09) | 34.51 (0.12) | -0.15 | **-0.43** | -0.50** | **-1.43** |
| Right upper lip | 35.09 (0.07) | 35.08 (0.12) | 34.72 (0.13) | -0.01** | **-0.03** | -0.37**ǂ | **-1.05** |
| Total face | 34.02 (0.04) | 33.73 (0.08) | 33.57 (0.04) | -0.28** | **-0.82** | -0.45** | **-1.32** |

Test of Normality whit Kolmogorov-Smirnov.
P-value with Student`s T for paired samples or ǂWilcoxon rank: *p < 0.05, **p < 0.001

Table 2 shows a dominant trend of temperature decrease for the negative valence emotion compared to the baseline in most cases and the ROIs and the p-value of statistics test was smaller than 0.001 and some case was less than 0.05, showing a statistically significant difference. There are some exceptions, for example, there is an increase of 0.05 °C in the case of the ROI of the left upper lip region of the man. This value is not significant because it is within the sensitivity range of the thermographic camera used, in addition the p-value of t-test was greater than 0.05 and therefore this change is not significative. Moreover, in the case of the woman 1, there is a temperature increase of 0.14 °C in the ROI of the right orbital, here the p-value is smaller than 0.001 showing a significative change. On the other hand, there are no thermal changes in the ROIs of the left orbital, the cheeks and the total face temperature, because the temperature differences are less than 0.05 °C and the p-value is greater than 0.05, which is the sensitivity value of the thermal thermographic camera used in this study.

Table 2 also compares the temperature for the positive valence emotion with the baseline. In the case of the man and the woman 1, there are several ROIs that do not show any significant thermal changes, where the temperature difference is less than 0.05 °C, which is within the sensitivity value of the thermographic camera used. The ROIs where there are no thermal changes are the forehead, right cheek, right and left upper lip for the man, and the nose, right cheek, and entire face for the woman 1. Note that the p values for these cases were smaller than 0.001, indicating a statistically significant differences, but is necessary considering both technical and statistical aspects, therefore, these temperature difference cannot be considered significative, since these small changes are statistical fluctuation of the measurement. However, for the woman 2, all ROIs show a temperature difference above 0.05 °C and p values smaller than 0.001, presenting a significant temperature difference.

To observe a thermal pattern in the face associated with the positive and negative valences, the thermal percentage differences of the ROIs were plotted ($\Delta T_{FB}$% and $\Delta T_{HB}$%). These are presented from the right to the left; the nose and the forehead are in the central part. Figure 8(a) shows the thermal pattern of the negative valence, and Figure 8(b) shows the thermal pattern of the positive valence.

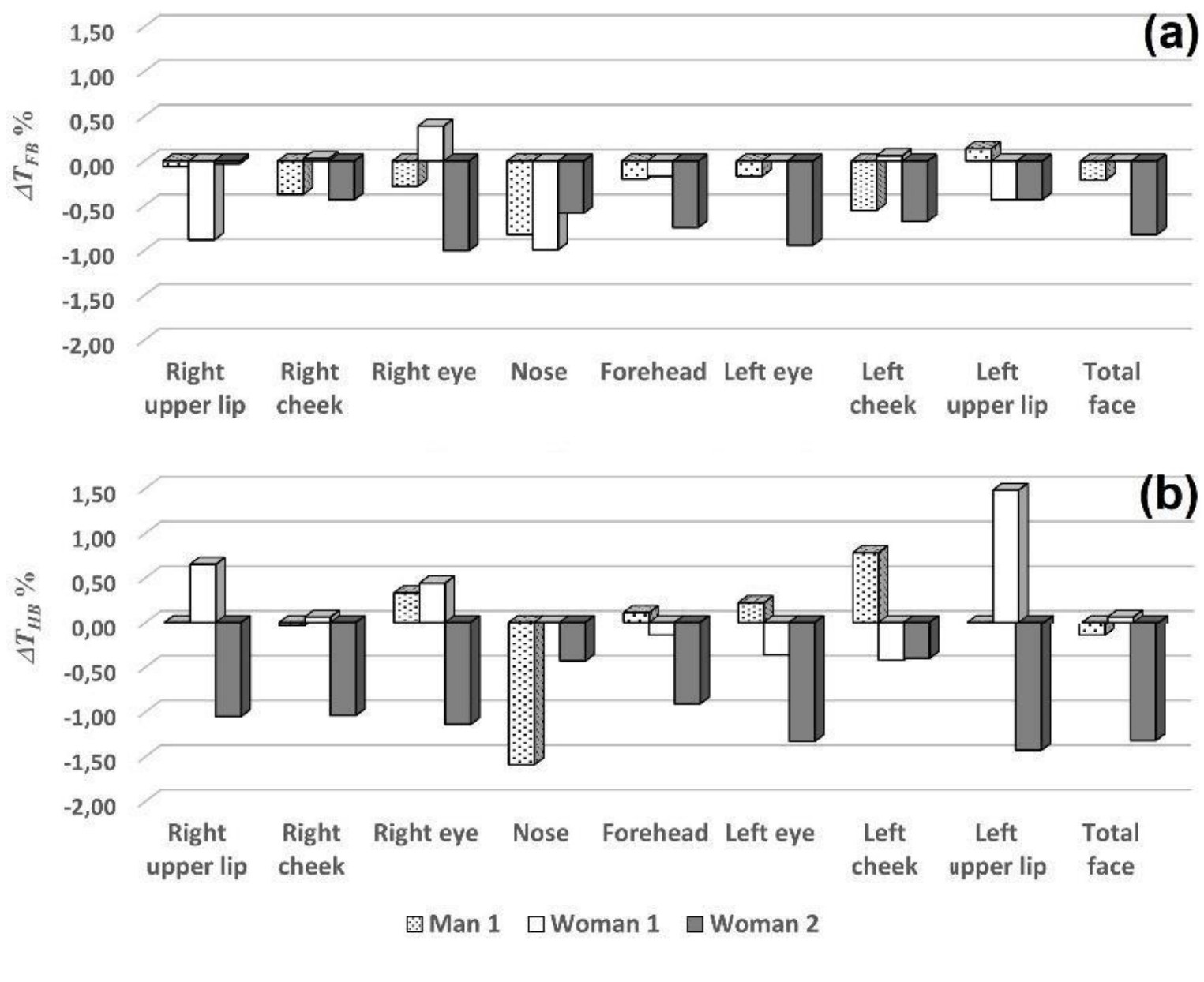

**Figure 8.** Values of the percentage differences between the average temperatures of the baseline and after applying the stimuli of (a) negative and (b) positive valence.

Figure 8(a) shows that the temperature of the nose and the temperature of the forehead decrease in the three analyzed individuals; further, on the right side of the face, the temperature decrease is greater than that on the left side, with some fluctuations. The foregoing shows the thermal behavior of the face as a response to the negative valence. Our results are consistent with what was reported in the work of Ioannou et al. [15], who indicated that the temperature is reduced in the nose, forehead, and right and left upper lip for the emotion of fear.

Figure 8(b) shows the thermal response of positive valence. It is observed that the temperature of the nose decreases in the analyzed ROIs. Temperatures may increase or decrease, for example, in the region of the right and left upper lip. There is no thermal change for the man. For woman 1, the temperature increases; for woman 2, the temperature decreases. In each point, the temperature may increase or decrease depending on the individual analyzed. The temperature variability of the face shows a response to the positive valence.

Fear and happiness are high arousal emotions (see Figure 1). Previous works [2], [21]–[23] have indicated that the nose's temperature decreases when the emotion is of a high arousal, like of fear and happiness. In this study, the nose's temperature decreases in the three subjects analyzed for both emotions. The absolute temperature variations ranged from −0.55 °C and −0.15 °C (see Table 2, Figures 8(a) and (b)), which present the emotions of fear and happiness evince that the nose's temperature decreases significantly. Our results confirm what was published in previous research [2], [21]–[23].

This study calculated the $\Delta T_{\alpha B}\%$, and it was found that the values are below 2% in all cases. The image was obtained in a controlled environment, which significantly reduces fluctuations in the temperature values, thereby ensuring that this percentage is due to a thermal change in response to the emotion studied.

### *3.2. Blood perfusion images analysis*

Table 3 presents the results of the average blood perfusion obtained for the baseline ($\omega_B$), the negative valence ($\omega_F$), and the positive valence ($\omega_H$) in the case of the man and the two women. Moreover, the values of the absolute differences ($\Delta\omega_{FB}$ and $\Delta\omega_{HB}$) and percentage differences ($\Delta\omega_{FB}\%$ and $\Delta\omega_{HB}\%$) are presented, which were calculated by means of Equations (3) and (4), respectively.

In Table 3, when comparing the blood perfusion values for the negative valence emotion with respect to the baseline, for the three individuals studied, a reduction in blood perfusion is observed except in the case of the right orbital region and left cheek of woman 1, whose percentage differences are below 1%, and for the man's right upper lip, where blood perfusion increases by 1.68%. In these cases, there are not statistically significant differences (p-values > 0.05).

Table 3 also compares the blood perfusion values for the positive valence emotion with the baseline. Here, a reduction in the blood perfusion of the nose region is observed for the three subjects studied. In the forehead, the right orbital region, and the right cheek are observed a decrease in the responses the blood perfusion of both women, and an increase in the man. In above cases, there are statistically significant differences (p-values < 0.001).

In the upper lip region, for woman 1, blood perfusion increases; for woman 2, it decreases. In the cases of both women, the percentage differences are above 7%, in addition to presenting statistically significant differences (p-values < 0.001). On the other hand, in the case of men, there is no change in the blood perfusion value in the upper lip region, since the percentage differences were insignificant, even one of them was zero and the other that gives a small difference was not presents statistically significant difference (left upper lip: $\Delta\omega_{HN}$ = 0.15, p-values > 0.05; right upper lip: $\Delta\omega_{HN}$ = 0.00, p-values < 0.05). This behavior of blood perfusion is the same as the one presented in the thermographic images.

**Table 3**. Average blood perfusion values in each region of interest for the baseline, negative valence, and positive valence and the absolute and percentage differences. All parameters are expressed in ×10$^{-2}$ [kg/m$^2$s], except for the percentages, which are nondimensional. Uncertainties are standard deviations, at 1 level, are given between parentheses.

| | $\omega_B$ **Baseline** | $\omega_F$ **Negative Valence** | $\omega_H$ **Positive Valence** | $\Delta\omega_{FN}$ | $\Delta\omega_{FB}$ **%** | $\Delta\omega_{HB}$ | $\Delta\omega_{HN}$ **%** |
|---|---|---|---|---|---|---|---|
| **Man 1** | | | | | | | |
| Noseǂ | 6.05 (0.28) | 5.53 (0.21) | 5.09 (0.28) | -0.53** | **-8.76** | -0.96** | **-15.87** |
| Forehead | 6.85 (0.05) | 6.69 (0.09) | 6.92 (0.04) | -0.16** | **-2.33** | 0.07** | **1.02** |
| Left eyeǂ | 9.47 (0.46) | 9.26 (0.47) | 9.77 (0.38) | -0.21 | **-2.17** | 0.30** | **3.16** |
| Right eyeǂ | 9.32 (0.95) | 8.98 (0.81) | 9.70 (0.43) | -0.34 | **-3.64** | 0.38** | **4.08** |
| Left cheek | 6.01 (0.23) | 5.66 (0.21) | 6.56 (0.13) | -0.35** | **-5.82** | 0.55* | **9.15** |
| Right cheek | 6.79 (0.15) | 6.51 (0.19) | 6.76 (0.06) | -0.28** | **-4.12** | -0.03** | **-0.44** |
| Left upper lip | 6.53 (0.13) | 6.64 (0.09) | 6.54 (0.18) | 0.11 | **1.68** | 0.01 | **0.15** |
| Right upper lip | 6.15 (0.06) | 6.10 (0.11) | 6.15 (0.06) | -0.05** | **-0.81** | 0.00** | **0.00** |
| Total face | 5.32 (0.03) | 5.19 (0.08) | 5.24 (0.03) | -0.13** | **-2.44** | -0.08** | **-1.50** |
| **Woman 1** | | | | | | | |
| Noseǂ | 5.60 (0.79) | 5.00(0.38) | 5.55 (0.36) | -0.60** | **-10.71** | -0.05 | **-0.89** |
| Forehead | 8.75 (0.14) | 8.55 (0.31) | 8.57 (0.13) | -0.20**ǂ | **-2.28** | -0.18** | **-2.06** |
| Left eye | 11.36 (1.71) | 11.13 (0.88) | 10.50 (0.33) | -0.23 | **-2.02** | -0.86 | **-7.57** |
| Right eyeǂ | 10.93 (3.09) | 10.98 (1.47) | 10.87 (0.74) | 0.05** | **0.46** | -0.06** | **-0.55** |
| Left cheek | 7.89 (0.37) | 7.92 (0.27) | 7.45 (0.17) | 0.03 | **0.38** | -0.44 | **-5.58** |
| Right cheekǂ | 8.57 (0.55) | 8.55 (0.30) | 8.63 (0.19) | -0.02 | **-0.23** | 0.06** | **0.70** |
| Left upper lip | 6.24 (0.35) | 5.94 (0.49) | 7.39 (0.22) | -0.30** | **-4.81** | 1.15**ǂ | **18.43** |
| Right upper lip | 7.63 (0.91) | 6.81 (0.54) | 8.18 (0.27) | -0.82** | **-10.75** | 0.55**ǂ | **7.21** |
| Total faceǂ | 5.72 (0.09) | 5.69 (0.16) | 5.75 (0.09) | -0.03 | **-0.52** | 0.03** | **0.52** |
| **Woman 2** | | | | | | | |
| Nose | 6.67 (0.52) | 6.63 (0.84) | 6.16 (0.49) | -0.04** | **-0.60** | -0.51**ǂ | **-7.64** |
| Forehead | 7.61 (0.15) | 6.92 (0.12) | 6.69 (0.06) | -0.69** | **-9.07** | -0.92** | **-12.09** |
| Left eyeǂ | 11.28 (0.49) | 9.65 (0.91) | 9.11 (0.81) | -1.63** | **-14.45** | -2.17** | **-19.24** |
| Right eyeǂ | 9.49 (1.26) | 8.27 (1.00) | 8.42 (1.26) | -1.22** | **-12.86** | -1.07** | **-11.28** |
| Left cheekǂ | 5.76 (0.39) | 5.27 (0.51) | 5.53 (0.52) | -0.49** | **-8.51** | -0.23** | **-3.99** |
| Right cheek | 7.19 (0.52) | 6.80 (0.52) | 6.20 (0.70) | -0.39** | **-5.42** | -0.99** | **-13.77** |
| Left upper lip | 7.33 (0.40) | 6.85 (0.28) | 6.12 (0.24) | -0.48 | **-6.55** | -1.21** | **-16.51** |
| Right upper lip | 7.45 (0.32) | 7.41 (0.27) | 6.55 (0.27) | -0.04** | **-0.53** | -0.90**ǂ | **-12.08** |
| Total face | 5.86 (0.10) | 5.40 (0.14) | 5.11 (0.05) | -0.46** | **-7.85** | -0.75** | **-12.80** |

Test of Normality whit Kolmogorov-Smirnov.
P-value with Student`s T for paired samples or ǂWilcoxon rank: *p < 0.05, **p < 0.001

In Figure 9(a), the percentage differences of blood perfusion between the negative valence and the baseline are presented. It is observed that most of the face ROIs for the three individuals studied present a reduction in blood perfusion. The percentage differences are greater on the left side than on the right side in most points. This result is consistent with the result obtained via the temperature analysis.

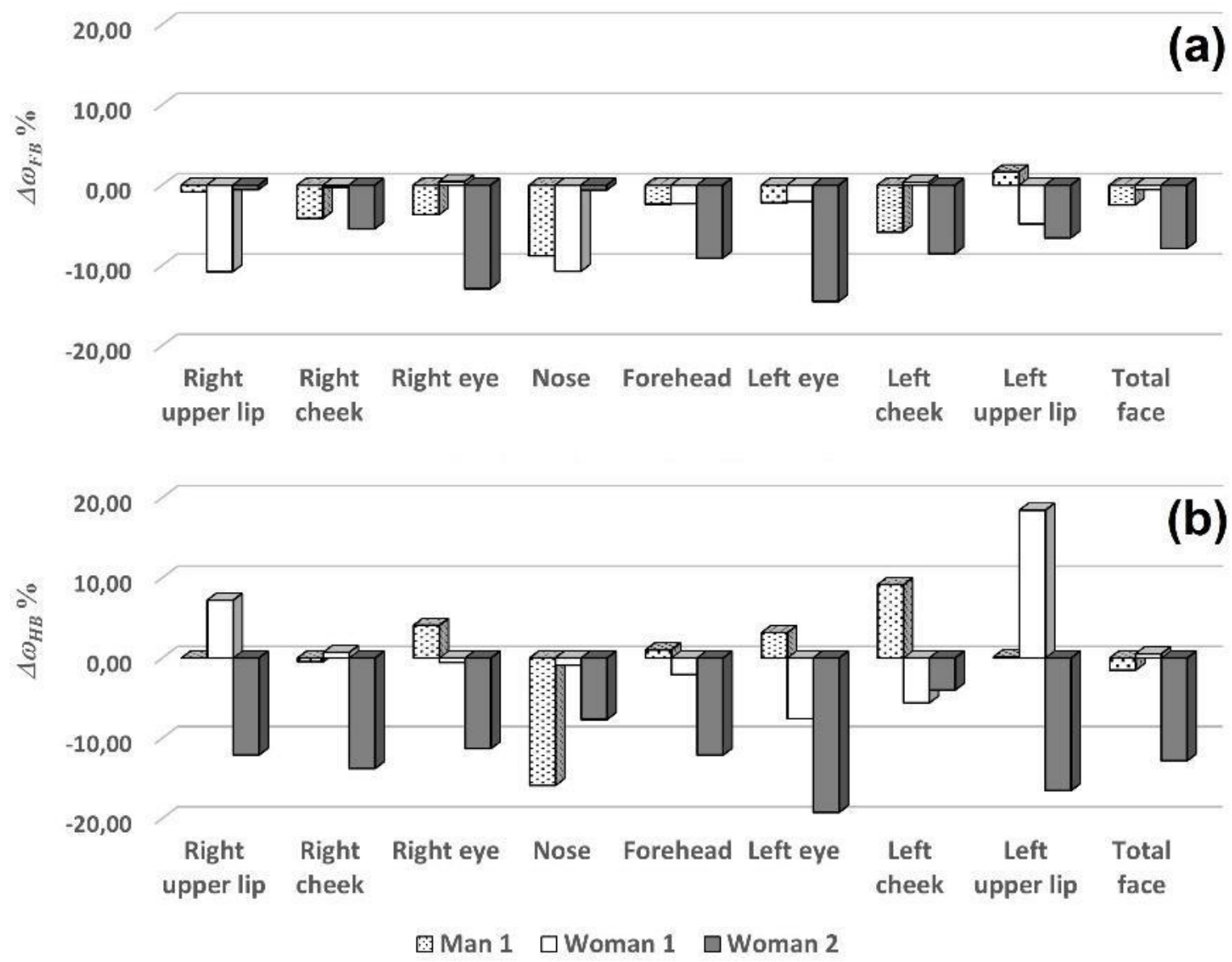

**Figure** 9. Values of the percentage differences between the average blood perfusion of the negative valence and the baseline.

Figure 9(b) shows the percentage differences in blood perfusion between the positive valence and the baseline. It is observed that the blood perfusion of the nose is reduced, as in the case of negative valence. In the other face ROIs, blood perfusion can be increased or decreased. These results are consistent with temperature analysis.

Note that the blood perfusion of the nose is reduced in both emotions, which makes sense because fear and happiness are emotions of high arousal. This same effect is observed in the thermographic images. The reduction in the blood perfusion of the nose can be used to measure the arousal dimension in individuals.

## 4. Discussion from a psychophysiological approach

By processing thermographic images with the blood perfusion algorithm, images that provide information about the blood flow in a person's face are obtained. Among the findings in this study, the differences in the perfusion of the right and left sides of some regions of the face stand out (Table 3). The responses of facial skin blood flow during exposure to a stimulus have great intersubject variation [48].

This physiological behavior could also be explained by few psychological mechanisms. The first is arousal [49], that is, when faced with negative valence stimuli, the activation of the organism is synchronized more consistently than when faced with positive valence stimuli [50]. The second mechanism is the emotional regulation [51], which is the mental activation of goals that modify the emotion. In negative valence stimuli, an additional effort is made to modify the unpleasant experience, which would require a greater effort and cognitive load, thereby making the physiological activation more evident [52]. Although the internal psychological mechanisms underlying this regulation remain unclear, it seems that orienting attentional networks would oversee leading this emotional processing [51].

Among the physiological mechanisms reported in the literature in relation to differences in facial perfusion are interhemispheric asymmetry and innervation of the face. The muscles of the face are predominantly innervated by the seventh cranial pair (facial nerve). The musculature of the upper part of the face is under bilateral cortical control, the two lower thirds of the face are contralaterally innervated. Consequently, from the lower eyelid down, the left side of the face is controlled by the right hemisphere, and the right side of the face is controlled by the left hemisphere. Owing to the emotional dominance of the right hemisphere, the contralateral innervation of the lower face results in greater expressiveness and emotional intensity on the left side [53]. The interhemispheric asymmetry in the emotional processing that is evidenced in this study supports the hypothesis of emotional regulation before unpleasant stimuli, as such asymmetry would be the facilitating element of emotional responses [54], [55]; when this asymmetry is not present, it is known that there is a dysfunctional processing of behavior and social cognition [56].

The other physiological mechanism is vascular given by the autonomic nervous system. In the case of an emotion of negative valence, the results of table 3 show a reduction of the blood flow in the total face that coincides with the physiological behavior that is due to aversive stimuli, which activate the sympathetic nervous system, increasing the cardiac output, blood pressure, and the redistribution of blood perfusion to vital organs. In the face, there is an increase in the diameter of the pupils to increase the visual field, the muscles contract, and vasoconstriction occurs, which generates an increase in resistance and a reduction in the blood flow [40]. A study by Mitja Benedičič et al. [57] reported asymmetry in the face's average blood flow and statistically significant differences in the spectral amplitude of the vasomotion (defined as spontaneous low-frequency oscillatory constrictions of the microvascular smooth muscle), which depend on intrinsic cardiac, respiratory, and myogenic activities and neural and endothelial mechanisms. The authors reported a higher degree of vasomotion activity on the right side of the forehead and a greater level of blood flow and vasomotion in the cheeks than on the forehead [57]. Additionally, the present findings are consistent with those reported by Matsukawa et al. [48], who found that the more pleasurable or conscious the emotional state becomes, the lower the blood flow is in the mouth, cheeks, and chin. The authors suggest, as an explanation for this decrease, the response given by the autonomic nervous system, which causes vasoconstriction and/or lack of vasodilation: vasoconstriction occurs because of the sympathetic stimulus that increases adrenaline, and vasodilation is due to a decrease in the activity of the parasympathetic system. This would support the vascular theory of emotional efference [58], [59], which posits the vascular facial system as a key element in processing emotions [60].

Finally, it is important to emphasize that the proposal of this work focuses on the need to have innovative tools to analyze the physiological aspects of the phenomenon of emotion, namely, tools that allow researchers to transcend the subjective measurement of this aspect. It has been reported that the internal consistency of emotion measurement with verbal instruments ranges from 0.61 to 0.85, while that of temporal stability is 0.81 [61]. These values remain stable in other languages as well, such as Turkish [62] and Spanish [63]. Such an innovation would present advantages in the evaluation of psychological aspects using techniques that are based on the measurements of physical variables associated with images. Another clear advantage is the language, context, and culture of the evaluation. To use verbal evaluation instruments, initial linguistic adaptation studies must be conducted [64]. The proposed technique goes beyond the linguistic limitations, as it only requires a thermographic camera to measure heat emission patterns, which is an objective process that is independent of the individual and the context. This leads to a consideration of the need for training on these types of techniques for professionals in the field of psychology. Although the findings here focus on emotional valence, further analyses could be conducted on discrete emotions (such as joy, sadness, and disgust, among others). In other physiological studies on such emotions, a relation was found between half the subjective elements; thus, whether this technique allows for a higher level of precision than classic physiological measurements, such as facial electromyography and electrodermic activity, could be investigated [65], [66]. Additionally, these findings in the future could be applied in neuropathologies [67], [68] and in psychiatric disorders such as depression or anorexia [69], [70].

**5. Conclusions**

The present work has evaluated, by a preliminary proof of concept study, the performance of the thermographic and blood perfusion data when emotions of positive and negative valence are applied. On the one hand, the images generated with the blood perfusion algorithm present higher contrast and greater number of qualitative details than thermographic images. On the other hand, when quantifying these values in the ROIs, percentage differences higher than 2% are observed and new results in the ROIs are obtained, which were not captured through thermographic images. In addition, the blood perfusion value presents less variability in the face of changes in environmental temperature and humidity, which implies that there is a reduction in noise due to environmental fluctuations in the measurements, finally, allowing greater stability in the identification of emotions in uncontrolled environments, as indicated in [25]. Therefore, the blood perfusion images have a potential to reach higher success rates in identifying emotions.

This study also shows that the blood perfusion changes can be a better biomarker to emotions than thermal changes; moreover, these blood perfusion patterns are consistent with the response to emotions from a psychophysiological approach. Additionally, the results of this work are supported by the vascular theory of emotional efference [60], as is emotional self-regulation of arousal due to consistency of physiological activation and interhemispheric brain asymmetry [55].

It is also important to emphasize that this type of study has note been conducted in Colombia, therefore, any contribution that can be made is considered of the utmost importance.

**6. Limitations and Future Lines of Research**

The main limitation of the study was the size of the sample, limiting the study to just three cases. A significant sample would allow us to quantify the range of variability of thermal and blood perfusion responses in the face, beside to compare with demographic variables as sex and age. It is necessary to analyze the significance of the differences among the emotional states and baseline by statistical tests according to the statistical distribution of data of a representative sample of the populations. Additionally, studies are required in uncontrolled conditions and with people with illnesses, to evaluate this proposal in natural environments and its effectiveness.

As this study focused on a dimensional approach of emotions, a future study would compare this novel analysis in a categorical approach of emotions (i.e., anger, sadness, happiness). In further studies we have considered validating and using videos for use in the Colombian population, also we consider necessary use self-report test as Self-Assessment Manikin (SAM). Additionally, a future study could implement the tracking procedure [71], [72] in the software image processing, which allows to identify the movement of face and to calculate the temperature and blood perfusion values with major accuracy in the different regions of interest and related blood perfusion to classical physiological measures as electrodermal activity (EDA) or heart rate variability (HRV).

This proposal could be applied in training of emotional skills in psychology students, this would be useful for emotional management in the interaction with patients in different performance contexts (i.e., schools, industries, etc.).

**Funding:** This research was funded by Universidad Cooperativa de Colombia, grant numbers INV2287 and INV1904.

**Institutional Review Board Statement:** The study was conducted according to the guidelines of the Declaration of Helsinki, and approved by the Ethics Committee of Universidad Cooperativa de Colombia, the Medellín campus where the study was conducted (Report N° 009).

**Informed Consent Statement:** Informed consent was obtained from all subjects involved in the study.

**Acknowledgments:** The authors would like to thank Universidad Nacional de Colombia, the Medellín campus, for the support given to obtain the Matlab licenses needed to develop the codes in this work. The authors would also like to thank Instituto Tecnológico Metropolitano for the lab equipment used for the thermographic measurements.

**Conflicts of Interest:** The authors declare no conflict of interest.

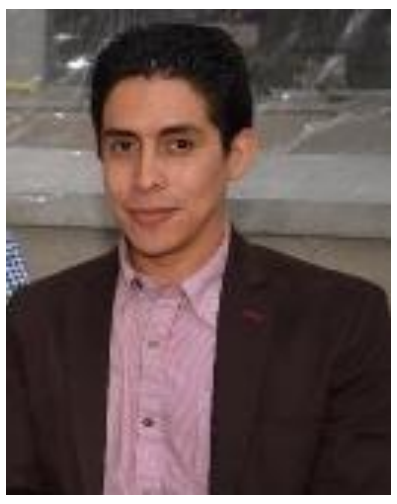

**Victor Hugo Aristizabal-Tique** obtained his B.S. degree in Physical Engineering from National University of Colombia in Medellin in 2004 and the M.Sc. in Physics in 2008 from the same university. He is currently doctoral student at Eafit University and full-time Research Professor at the Cooperative University of Colombia, Medellín, Colombia. His work has been focused on analytical and computational modelling of the elastic waves in seismic engineering and geophysical prospecting, and electromagnetic waves in optical communications and sensing.
ORCID: 0000-0002-7880-58839

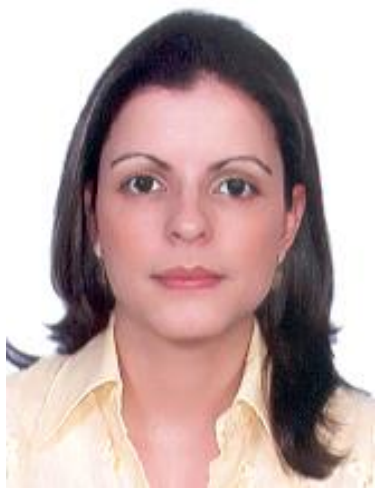

**Marcela Henao-Pérez** is physician from Universidad Pontificia Bolivariana, Medellín, Colombia (2002), magister in neurosciencie from Universidad Pablo de Olavide, Sevilla, España (2010), and Neuroscience Doctor Candidate from the same university. She is a fulltime Professor-Researcher at the Universidad Cooperativa de Colombia, medical school of Medellín, Colombia. Her research has focused in stress medicine, mental health, fibromyalgia, cardiovascular risk, and sepsis in prehospital medical care.
ORCID: 0000-0002-7337-2871

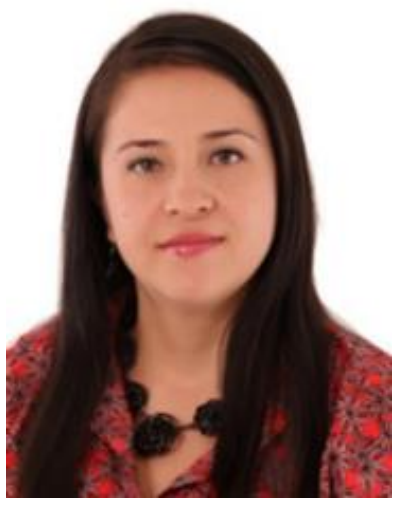

**Diana Carolina López-Medina** obtained her B.S. Graduated in Medicine from the CES University of Colombia in Medellín in 2007, the Esp. in Epidemiology from the CES University in 2016 and the Master in Epidemiology in 2018 from the same university. She is a fulltime Professor-Researcher at the Universidad Cooperativa de Colombia, medical school of Medellín, Colombia. Her work has focused on epidemiological studies on clinical issues such as cardiovascular risk, mental health, sepsis in prehospital medical care, fibromyalgia.
ORCID: 0000-0003-2098-7319

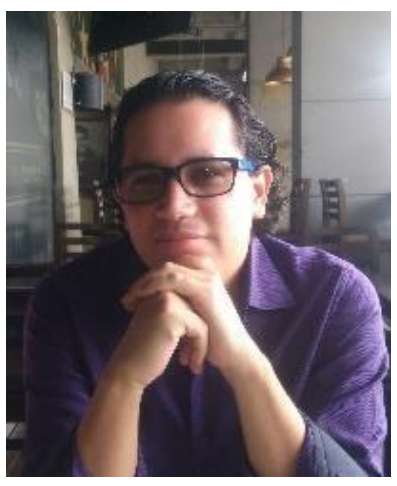

**Renato Zambrano-Cruz** is psychologist (2007) and magister in linguistics (2009) from University of Antioquia in Medellin, Colombia. He is doctor magna cum laude in cognitive sciences (2021) from Autonomous University of Manizales, Colombia. He has been working as assistant professor at Cooperative University of Colombia, campus Medellín. His fields of research are psychometrics, language, cognition, and emotions.
ORCID: 0000-0003-2155-0039

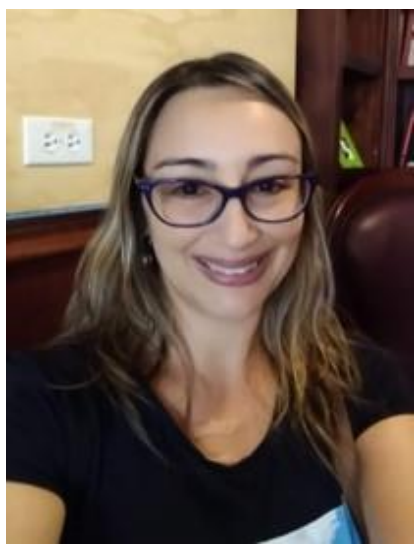

**Gloria Díaz-Londoño** received the B.S degree in Electrical Engineering in 2000 from National University of Colombia in Medellin, the M.Sc. in Physics in 2006 from the same University and PhD degree in Bioengeneering and Medical Physics in 2015 from Granada University, España. She is currently an associate professor with the School of Physics, National University of Colombia in Medellin. Her fields of research are medical Image processing, Infrared Thermography, radiation physics, dosimetry in nuclear medicine and Monte Carlo simulation.
ORCID: 0000-0002-3235-1193